%% file: main.tex
\PassOptionsToPackage{table}{xcolor}
\documentclass[letterpaper]{article} 
\usepackage[preprint]{aaai2027}
\usepackage[hyphens]{url}  
\usepackage{graphicx} 
\usepackage{natbib}  
\usepackage{caption} 
\usepackage{algorithm}

\usepackage{newfloat}
\usepackage{listings}
\DeclareCaptionStyle{ruled}{labelfont=normalfont,labelsep=colon,strut=off} 
\floatstyle{ruled}
\newfloat{listing}{tb}{lst}{}
\floatname{listing}{Listing}

\usepackage{booktabs}

\usepackage{amsmath}
\usepackage{amssymb}
\usepackage{algpseudocode}
\algrenewcommand\algorithmicrequire{\textbf{Input:}}
\algrenewcommand\algorithmicensure{\textbf{Output:}}

\colorlet{softgreen}{green!50!gray}
\usepackage{color}
\newcommand{\tzlconfirmed}[1]{{#1}}
\usepackage{multirow}
\usepackage{siunitx}
\usepackage[table]{xcolor}
\usepackage{adjustbox}
\usepackage{array}
\usepackage{tabularx}

\definecolor{oursgray}{gray}{0.92}

\newcolumntype{Y}{
    >{\centering\arraybackslash}X
}

\newcolumntype{W}{
    >{\hsize=1.20\hsize
      \linewidth=\hsize
      \centering\arraybackslash}X
}

\newcolumntype{N}{
    >{\hsize=0.90\hsize
      \linewidth=\hsize
      \centering\arraybackslash}X
}

\title{Beyond Query Similarity: Safety-Conditioned Cache Routing \\ for LLM Response Reuse}
\author{
    Shilong Pan\textsuperscript{\rm 1},
    Zhiliang Tian\textsuperscript{\rm 1}\corresponding,
    Weiguo Chen\textsuperscript{\rm 1},
    Zihan Chen\textsuperscript{\rm 1},
    DongSheng Li\textsuperscript{\rm 1}\corresponding,
}
\affiliations{
    \textsuperscript{\rm 1}PDL Lab, College of Computer Science and Technology\\
    National University of Defense Technology

    panshilong18@nudt.edu.cn
}

\begin{document}

\maketitle

\begin{abstract}

\input{sections/abstract}

\end{abstract}


\input{sections/introduction}

\input{sections/related_work}

\input{sections/methodology}
\input{sections/experiments}

\input{sections/conclusion}

\bibliography{aaai2027}

\clearpage
\appendix
\input{sections/appendix_v5}


\end{document}

%% file: sections/abstract.tex
Semantic response caching reuses a cached response for a semantically similar query to improve LLM inference efficiency. However, query similarity alone can lead to unsafe reuse when safety conditions change. Existing approaches improve cache safety through stricter similarity checks, context-aware validation, or response verification, yet they may still miss safety conflicts between the cached response and the new request or introduce unnecessary verification overhead.
We argue that safe and efficient semantic caching requires a reuse policy that makes the cached response's safety conditions explicit and uses them to guide the reuse decision with a balance between safety and utility.
In this paper, we propose SafeCache, a safety-aware framework that separates candidate retrieval from safety approval and formulates semantic response reuse as a safety-conditioned routing problem.
%
SafeCache matches each cached-response condition against the new request and aggregates the per-condition mismatches into an overall risk score. 
A separate contradiction guard prevents explicit conflicts from being diluted by strong matches.
It further estimates matching uncertainty and calibrates the routing policy under a required unsafe-reuse target and a verifier-call budget to choose among direct reuse, selective verification, or regeneration. Experiments show that SafeCache achieves the better safety-utility trade-off among the evaluated baselines, enabling safer response reuse at comparable cache utility.

%% file: sections/introduction.tex
\section{Introduction}

\tzlconfirmed{LLM services often receive repeated or semantically similar requests. 
Semantic caching reduces the inference time on repeated or similar requests by storing query-response pairs and reusing a cached response for a semantically similar new request \citep{gptcache2023,gptsemanticcache2024,meancache2025,contextcache2025}, which can reduce serving latency and inference cost \citep{scalm2024,lowcostsemanticcache2026}. 
However, similarity between queries does not guarantee that the cached response remains safe for the new request. 
Changes in safety-relevant conditions, such as the protected subject, authorization state, supporting evidence, or tool context, may invalidate the cached response. 
Recent studies further show that semantic caches can be exploited to return unsafe cached responses \citep{cacheattack2026,cachemecatchyou2026,cachepoisoning2026}. 
Therefore, semantic cache safety requires determining whether a cached response can be safely reused for each new request.}

\tzlconfirmed{Existing work addresses semantic cache safety in two main ways. The first group tries to make reuse safer through stricter similarity matching \citep{gptcache2023,gptsemanticcache2024,meancache2025,vcache2025}. These methods narrow the cache hit set by requiring the new query to be very close to a cached query. This can filter out obvious mismatches, but it still only compares the two queries and does not verify whether the cached response is actually safe to reuse, even when the queries appear semantically similar \citep{cacheattack2026,cachemecatchyou2026}.
The second group, context-aware caches, therefore, adds external context, metadata, or retrieved support evidence to the reuse decision \citep{contextcache2025,groundedcache2026}.  These checks make the cache safer when a decisive change appears in the external context. 
%
%
However, the external context may remain consistent even when the cached response is internally unsafe for the new request \citep{contextcache2025,groundedcache2026}. As a result, unsafe cached responses may still be reused.
}

\tzlconfirmed{To examine whether a cached response remains safe for a semantically similar new request, verified semantic caches add a verifier before returning a candidate response \citep{vcache2025,krites2026,groundedcache2026}. A verifier can check response safety conditions, including the protected subject, authorization state, evidence support, and tool state, rather than relying only on query similarity or limited external context. 
However, existing verified caches often treat verification as a generic approval step after retrieval. They do not clearly specify what safety conditions in the cached response should be checked, what types of safety conflicts should require response regeneration, or which uncertain candidates are worth the extra cost of a verifier.}
%
%
Consequently, they may reuse unsafe responses, undermining safety. They may also waste verifier calls on clear cases or discard responses that could be safely reused, reducing utility.
These limitations motivate a safety-conditioned reuse policy that explicitly represents response safety conditions and adapts the reuse decision to better balance safety and utility in semantic caching.


\tzlconfirmed{To better balance safety and utility, we argue that a safety-conditioned semantic cache needs a routing policy that makes the cached response's safety conditions explicit and uses them to balance safety and utility.} Similarity should only retrieve a candidate, while safety approval should depend on whether these conditions remain satisfied for the new request, allowing efficient reuse without undermining safety.

In this paper, we propose SafeCache, a safety-aware framework for semantic response caching that makes reuse decisions based on safety conditions, aiming to improve safe reuse while preserving caching utility.
\tzlconfirmed{Specifically, SafeCache first uses embedding similarity only to retrieve a plausible candidate, separating candidate retrieval from the final reuse decision. A safety-conditioned router then decides whether the cached response should be reused, verified, or regenerated. To capture safety changes that ordinary query similarity cannot detect, SafeCache extracts the safety conditions required by the cached response and matches them against those in the new request.} 
%
%
This condition-level matching yields an aggregated risk score that measures the overall mismatch between the new request and the safety conditions required for reuse.
To ensure that explicit safety conflicts are not overlooked, SafeCache checks them separately from the aggregated risk score and prevents reuse when such a conflict is detected.
\tzlconfirmed{Because verification adds extra cost, SafeCache further uses an uncertainty-aware safety matching strategy to identify borderline candidates and sends them to a verifier only when they appear reusable but are not reliable enough for direct reuse.
Extensive experiments show that SafeCache achieves a better safety-utility trade-off among the semantic cache baselines, with stronger control of unsafe response reuse under comparable cache utility.}

\tzlconfirmed{We make three contributions.
\begin{itemize}
\item We present SafeCache, a safety-aware framework for semantic response caching that separates candidate retrieval from the final reuse decision and explicitly targets the safety-utility trade-off in cache reuse.
\item We develop a guarded and uncertainty-aware safety routing algorithm that checks whether the cached response's safety conditions still hold, blocks direct reuse under explicit contradictions, and uses calibrated uncertainty to decide when to reuse, verify, or regenerate.
\item We introduce a semantic cache safety evaluation and show that SafeCache achieves a better safety-utility trade-off among the evaluated semantic cache baselines.
\end{itemize}}


%% file: sections/related_work.tex
\section{Related Work}

\subsection{LLM Caching}

For LLM text generation, caching systems reuse either model states associated with repeated token sequences or complete responses to semantically similar requests.
\noindent\textbf{(1) Prefix and KV Caching.}
Prefix and KV caches reuse attention states through modular prompt schemas, radix-tree prefix sharing, or memory-tiered storage \citep{promptcache2024,sglang2024,lmcache2025,vllm2023,hydragen2024,cachegen2024}. Other systems extend prefix reuse to agent workflows and hybrid model architectures \citep{kvflow2025,marconi2025}. These methods reduce repeated prefill computation but still depend on shared token sequences \citep{ragcache2024,cacheblend2025,turborag2024}.
\noindent\textbf{(2) Semantic Response Caching.}
Semantic caches store query-response pairs and retrieve prior answers through embedding similarity. GPTCache provides a configurable framework for this form of reuse, and subsequent work studies its latency, cost, and embedding-space cache policies \cite{gptsemanticcache2024,scalm2024,lowcostsemanticcache2026,biton2026closeenough,atalar2026continuous}. Later systems adapt semantic matching to personalized and multi-turn requests \citep{meancache2025,contextcache2025}. This broader reuse requires deciding whether a response generated for one request remains valid for another.

\subsection{Risks in LLM Caching}

Risks introduced by LLM caching can be divided into reuse validity risks and adversarial integrity risks.
\noindent\textbf{(1) Reuse Validity Risks.}
\citet{contextcache2025} shows that a cache hit can become invalid when dialogue context changes, even if the current queries remain similar. \citet{groundedcache2026} identifies changes in retrieved evidence and source versions as another source of invalid cached answers. Caches that rely on similarity thresholds also face uneven error rates across prompts, and strong ranking performance does not guarantee reliable decisions at a fixed threshold \citep{vcache2025,calibrationgap2026}.
\noindent\textbf{(2) Adversarial Integrity Risks.}
\citet{cacheattack2026} models semantic cache keys as fuzzy hashes and constructs black-box collisions that redirect hits to unintended responses. Cache poisoning inserts crafted query-response pairs so that later related queries retrieve content selected by the attacker \citep{cachepoisoning2026}. \citet{cachemecatchyou2026} analyzes attacks against users and the integrity of LLM serving systems across prefix and semantic caches.

\subsection{Safety Approaches for LLM Caching}

Existing safety approaches for LLM caching mainly fall into three categories.
\noindent\textbf{(1) Similarity-based Filtering.}
These methods accept cached responses using embedding similarity and predefined thresholds \citep{gptcache2023,gptsemanticcache2024,meancache2025}. 
\citet{vcache2025} further estimates a prompt-specific threshold under a target error bound, while calibration studies examine whether similarity scores support reliable threshold decisions \citep{calibrationgap2026}.
\noindent\textbf{(2) Context-aware Validation.}
These methods incorporate dialogue history, retrieved evidence, or source state into the reuse decision. ContextCache reranks candidates using current and historical dialogue \citep{contextcache2025}, whereas \citet{groundedcache2026} checks evidence overlap, source versions, and fresh retrieval support.
\noindent\textbf{(3) Verifier-based Validation.}
These methods inspect a retrieved response before returning it. Existing designs use LLM-based filters or coherence checks to reject misleading or poisoned cache hits \citep{cachemecatchyou2026,cachepoisoning2026}, and extend verification to cached knowledge or intermediate generation steps \citep{cortex2026,stepcache2026}. 

%% file: sections/methodology.tex
\input{figures/main_fig}

\section{Methodology}
\label{sec:methodology}

\subsection{Overview}
\label{sec:method-overview}

\tzlconfirmed{Our proposed SafeCache, as shown in Fig.~\ref{fig:main_fig}, consists of three parts. The first is Response Safety Modeling, which extracts separate safety condition sets} for the cached response and the new request. 
%
%
\tzlconfirmed{The second is Guarded and Uncertainty-Aware Safety Matching, which uses condition-level semantic matching to measure the degree to which the} cached response conditions remain supported by the new request, \tzlconfirmed{while separately checking for explicit contradictions and estimating the reliability of the matching result.}
%
%
%
%
The third is Safety-Conditioned Cache Routing, which combines retrieval similarity with calibrated safety evidence to make a cache routing policy under safety and verifier-call constraints.


Overall, SafeCache adopts a three-stage cache-routing framework. It first constructs separate safety condition sets for the cached response and the new request, and then evaluates their safety compatibility to derive mismatch risk and matching uncertainty. Finally, a calibrated router combines safety evaluation with query similarity to determine a cache reuse policy under safety and utility constraints.

\subsection{Response Safety Condition Modeling}
\label{sec:safety-condition-modeling}

To make response safety explicit before cache reuse, SafeCache constructs separate safety condition sets for the cached response and the new request. The motivation is that query similarity only indicates that two requests have similar context, whereas safe reuse also depends on the protected subject, authorization state, supporting evidence, tool state, and so on. Because these conditions may change even when the requests remain semantically similar, SafeCache models them explicitly before making the reuse decision.

SafeCache takes a cached entry $(q_c,r_c,c_c)$, where $q_c$ is the cached request, $r_c$ is the cached response, and $c_c$ is the associated context, together with a new request-context pair $(q_n,c_n)$, where $q_n$ is the new request and $c_n$ is its context. The cache-side extractor produces the cached-response condition set $\mathcal{E}_c=\operatorname{Extract}_{\mathrm{cache}}(q_c,r_c,c_c)$, which records the conditions required to reuse $r_c$ safely. The request-side extractor produces the new-request condition set $\mathcal{C}_n=\operatorname{Extract}_{\mathrm{request}}(q_n,c_n)$, which records the safety-relevant facts and states in $q_n$ and $c_n$. 
Detailed extraction rules are provided in App.~\ref{app:condition-extraction}.
By separating cached response requirements and new-request facts, this modeling enables detection of unsupported conditions and safety conflicts.
The next module compares $\mathcal{E}_c$ and $\mathcal{C}_n$ to determine whether the cached response remains safe to reuse.

\subsection{Guarded and Uncertainty-Aware Safety Matching}
\label{sec:safety-matching}


To assess whether the safety conditions for the cached response remain supported by the new request, SafeCache combines aggregate safety condition matching with explicit contradiction checks and matching uncertainty estimation. 
The motivation is that safe reuse depends on the joint support of multiple cached-response conditions, which SafeCache summarizes with an aggregate mismatch score. Because the aggregate score may hide explicit conflicts or be unreliable under ambiguous matches, SafeCache adds explicit contradiction checks and a separate uncertainty estimate.

\subsubsection{Aggregate Condition Support Matching}

To obtain an overall measure of safety compatibility, SafeCache matches each cached-response condition against the new-request condition set and aggregates the resulting per-condition mismatches.
\tzlconfirmed{SafeCache examines each condition in $\mathcal{E}_c$ one at a time, treating it as a safety requirement, and searches $\mathcal{C}_n$ for the condition that provides the strongest support. The resulting per-condition mismatches are then averaged into a continuous overall mismatch score. }


\tzlconfirmed{Let $M$ and $N$ denote the numbers of conditions in the cached-response set $\mathcal{E}_c$ and the new-request set $\mathcal{C}_n$, respectively. For each cached-response condition $e_i$, SafeCache compares it with every new-request condition $c_j$ using the similarity calculation $\operatorname{sim}_{\phi}$, where $\phi$ is the frozen condition representation model.}
%
%
%
SafeCache retains the highest pairwise score $m_i^{(1)}$ as the strongest support in $\mathcal{C}_n$ for $e_i$. The per-condition mismatch is then defined as $1-m_i^{(1)}$, so weaker support yields a larger mismatch.
\tzlconfirmed{SafeCache then averages the per-condition mismatches to keep the aggregate on the same scale when the number of cached-response conditions varies. This produces the raw condition mismatch $\overline{D}_{\mathrm{match}}$ as}
\begin{equation}
\begin{aligned}
m_i^{(1)}
&=\max_{1\leq j\leq N}
\operatorname{sim}_{\phi}(e_i,c_j),\\
\overline{D}_{\mathrm{match}}
&=\frac{1}{M}\sum_{i=1}^{M}
\left(1-m_i^{(1)}\right).
\end{aligned}
\label{eq:condition-matching}
\end{equation}


Because each pairwise similarity lies in $[0,1]$, the raw condition mismatch also lies in $[0,1]$ and increases as more cached-response conditions receive weak or no support. 
By aggregating condition-level mismatch scores, this score provides a normalized and continuous measure of overall safety compatibility, capturing different degrees of support.

\subsubsection{Explicit Contradiction Checks}

%
%
To prevent decisive safety conflicts from being diluted by aggregate matching, SafeCache detects explicit contradictions as an independent signal against direct reuse. It checks both extracted condition records and original inputs, in which the former captures structured conflicts, while the latter covers conflicts missed during extraction.
\tzlconfirmed{The two checks are combined into an independent conflict indicator that blocks direct reuse whenever an explicit contradiction is detected.}

At the condition level, SafeCache applies conflict rules $\mathcal{R}_{\mathrm{rule}}$ (App.~\ref{app:condition-extraction}) to the condition sets of cached response and new request (i.e., $\mathcal{E}_c$ and $\mathcal{C}_n$). 
\tzlconfirmed{Each rule $r\in\mathcal{R}_{\mathrm{rule}}$ first determines whether two conditions concern the same safety decision and then checks whether their values, states, or required response actions conflict. The resulting indicator $H_{\mathrm{rule}}(x)=1$ whenever at least one condition-level rule detects an explicit contradiction: $H_{\mathrm{rule}}(x)=\bigvee_{r\in\mathcal{R}_{\mathrm{rule}}}r(\mathcal{E}_c,\mathcal{C}_n)$, where $x=(q_c,r_c,c_c,q_n,c_n)$ denote the cache candidate and each rule $r(\cdot)$ returns a binary value.}

At the input level, SafeCache applies conflict rules $\mathcal{R}_{\mathrm{ctx}}$
directly to the cached and new request-context pairs $(q_c,c_c)$ and $(q_n,c_n)$. These rules compare corresponding safety-relevant attributes, such as authorization state, tool outcome, and requested time, to detect explicit conflicts that may not have been preserved during condition extraction. 
\tzlconfirmed{The resulting indicator $H_{\mathrm{ctx}}(x)=1$ whenever any input-level rule detects such a conflict: $H_{\mathrm{ctx}}(x)=\bigvee_{r\in\mathcal{R}_{\mathrm{ctx}}}r(q_c,c_c,q_n,c_n)$.}

\tzlconfirmed{As a result, SafeCache retains $H_{\mathrm{rule}}$ and $H_{\mathrm{ctx}}$ separately to record the level at which a conflict is detected, and combines them into the final conflict indicator,
\begin{equation}
    H(x)=H_{\mathrm{rule}}(x)\lor H_{\mathrm{ctx}}(x).
\label{eq:explicit-conflict-indicator}
\end{equation}}
%
%
%
%
Thus, $H(x){=}1$ whenever either level detects an explicit contradiction. Unlike the aggregate condition mismatch, $H$ serves as an independent Boolean guard that cannot be diluted by averaging in Eq.~\eqref{eq:condition-matching} and directly blocks reuse.

\subsubsection{Uncertainty-Aware Risk Estimation}

\tzlconfirmed{To assess the reliability of condition matching, SafeCache estimates matching uncertainty $U_{\mathrm{match}}$ by measuring how clearly each cached-response condition is distinguished from alternative matches. The raw mismatch in Eq.~\eqref{eq:condition-matching} uses the strongest support score $m_i^{(1)}$ but does not reflect whether other candidates receive similar scores. When several candidates have similar scores, the selected match is less stable.}

\tzlconfirmed{For each cached-response condition $e_i$, SafeCache retains the strongest score $m_i^{(1)}$ and the second-highest score $m_i^{(2)}$.
Their difference $m_i^{(1)}-m_i^{(2)}$ measures the confidence of the best correspondence \citep{joshi2009margin}.}
%
%
A large margin therefore indicates both strong support and a clearly distinguished match, while the margin is small when the best score is low or when the two highest scores are close.
\tzlconfirmed{SafeCache averages these margins across the cached-response conditions and converts the result into the matching uncertainty:}
\begin{equation}
\begin{aligned}
U_{\mathrm{match}}(x)
&=1-\frac{1}{M}\sum_{i=1}^{M}(m_i^{(1)}-m_i^{(2)}).
\end{aligned}
\label{eq:match-uncertainty}
\end{equation}
\tzlconfirmed{Because $0\leq m_i^{(2)}\leq m_i^{(1)}\leq1$, the matching uncertainty lies in $[0,1]$ and increases as the best matches $m_i^{(1)}$ become weaker or less distinct.} 
%
%

This matching module summarizes each candidate using three complementary signals: $\overline{D}_{\mathrm{match}}(x)$ measures overall condition mismatch, $U_{\mathrm{match}}(x)$ reflects the reliability of the matching result, and $H(x)$ identifies explicit safety conflicts. The following routing module then calibrates and combines these signals for the final cache decision.

\subsection{Safety-Conditioned Cache Routing}
\label{sec:safety-routing}

\tzlconfirmed{To convert a retrieved candidate into a cache decision, SafeCache combines query similarity with response-level safety evidence to choose direct reuse, verification, or regeneration. Query similarity indicates whether the cached request remains relevant, but does not determine whether the response is safe to reuse or whether verification is necessary. SafeCache therefore calibrates the matching risk and routing parameters on a separate calibration set and keeps them fixed during the test. Clear candidates are directly reused, selected borderline candidates are verified, and the remaining candidates are regenerated.}

\tzlconfirmed{For the cache candidate $x=(q_c,r_c,c_c,q_n,c_n)$, the retriever computes the query similarity $S(x)=\operatorname{sim}_{\psi}(q_c,q_n),$
where $\psi$ is the query encoder. The router uses $S$ only to measure whether the cached request remains relevant to the new request. The preceding matching module provides the raw condition mismatch $\overline{D}_{\mathrm{match}}$ in Eq.~\eqref{eq:condition-matching}, explicit conflict indicator $H$ in Eq.~\eqref{eq:explicit-conflict-indicator}, and matching uncertainty $U_{\mathrm{match}}$ in Eq.~\eqref{eq:match-uncertainty}. }
%
%
Before routing, SafeCache converts the raw matching outputs into a calibrated risk score $D$ and an uncertainty score $U$. The router then uses $S$, $D$, $U$, and $H$ to make the final cache decision.

\subsubsection{Risk Calibration}

\tzlconfirmed{The raw condition mismatch $\overline{D}_{\mathrm{match}}$ in Eq.~\eqref{eq:condition-matching} measures how poorly the new request supports the cached-response conditions, but it does not directly represent unsafe-reuse risk \citep{guo2017calibration}. SafeCache therefore learns a monotonic calibration function $\mathcal{K}$ from a labeled calibration set to map mismatch scores to observed unsafe-reuse rates, ensuring that a larger mismatch cannot produce a lower calibrated risk.}

\tzlconfirmed{Because each calibrated risk is estimated from a limited number of examples, SafeCache also measures its uncertainty. It uses the width of the corresponding Wilson interval \citep{wilson1927probable} as $U_{\mathrm{cal}}(x)\in[0,1]$, in which a narrower interval indicates a more reliable estimate, while a wider interval indicates greater uncertainty. Further details are in the App~\ref{app:risk-calibration}.}

\tzlconfirmed{SafeCache then uses $U_{\mathrm{cal}}(x)$ to balance the calibrated risk and the raw mismatch. When calibration uncertainty is low, the final routing risk relies more on $\mathcal{K}(\overline{D}_{\mathrm{match}}(x))$. As uncertainty increases, it retains more of $\overline{D}_{\mathrm{match}}(x)$. Accordingly, the final routing risk score is defined as
\(
D(x)
=
\bigl(1-U_{\mathrm{cal}}(x)\bigr)
\mathcal{K}\!\left(\overline{D}_{\mathrm{match}}(x)\right)
+
U_{\mathrm{cal}}(x)
\overline{D}_{\mathrm{match}}(x).
\)}

The final uncertainty $U(x)$ accounts for uncertainty in both condition matching and risk calibration. SafeCache defines $U(x)
\!=\!
\max\!\left\{
U_{\mathrm{match}}(x),
U_{\mathrm{cal}}(x)
\right\}$, to preserve high uncertainty in either step.
The resulting $D(x)$ and $U(x)$ are used as the risk and uncertainty signals for routing.

\subsubsection{Three-Way Cache Routing}


SafeCache uses a calibrated three-way policy $\Theta$ to route each candidate $x$ to direct reuse, selective verification, or regeneration based on query similarity $S(x)$, reuse risk $D(x)$, uncertainty $U(x)$, and explicit safety conflicts $H(x)$.

\noindent\textbf{Direct reuse.}
Let $\tau_r$ denote the minimum query similarity for direct reuse, and let $\delta_r$ and \tzlconfirmed{$\upsilon_r$ denote the maximum allowed risk and uncertainty, respectively. A candidate is directly reused only if the cached request remains sufficiently relevant, $S(x)\geq\tau_r$; its calibrated risk and uncertainty are sufficiently low, $D(x)\leq\delta_r$ and $U(x)\leq\upsilon_r$; and no explicit contradiction is detected, $H(x)=0$. These conditions define the direct reuse region
\begin{equation}
\mathcal{R}_{\Theta}
\!=\!
\left\{
x\ \middle|
H(x)\!=\!0,
S(x)\!\geq\!\tau_r,
D(x)\!\leq\!\delta_r,
U(x)\!\leq\!\upsilon_r
\right\}.
\label{eq:reuse-region}
\end{equation}
Candidates in $\mathcal{R}_{\Theta}$ return the cached response directly.}

\noindent\textbf{Selective verification.}
Candidates rejected by the direct-reuse test may still be worth verifying when they remain relevant to the new request.  
Let $\tau_v$ denote the minimum similarity for verification, and $\delta_v$ and $\upsilon_v$ the maximum allowed risk and uncertainty. Compared with direct reuse, verification uses a lower similarity threshold ($\tau_v \leq \tau_r$) and higher limits on risk ($\delta_v \geq \delta_r$) and uncertainty ($\upsilon_v \geq \upsilon_r$), allowing less reliable but still relevant candidates to be verified.

An explicit conflict $H(x)\!=\!1$ blocks direct reuse but may still be resolved by verification when the cached and new requests are sufficiently related (see App.~\ref{app:signal-diagnostics} for diagnostic analysis). SafeCache therefore applies a stricter similarity threshold $\tau_v^{c}\geq\tau_v$ to conflict candidates. The verification threshold for each case is defined as
\begin{equation}
\widetilde{\tau}_v(x)=
\begin{cases}
\tau_v^{c}, & \text{if } H(x)=1,\\
\tau_v, & \text{if } H(x)=0.
\end{cases}
\label{eq:verify-threshold}
\end{equation}
The condition $S(x)\geq\widetilde{\tau}_v(x)$ provides the similarity requirement for verification.

The router then combines the similarity requirement with the verification limits of risk $\delta_v$ and uncertainty $\upsilon_v$, which require the candidate risk and uncertainty to remain within the range admitted to verification:
\begin{equation}
\begin{aligned}
\mathcal{V}_{\Theta}
\!=\!\left\{
x \,\middle|\,
x\notin\mathcal{R}_{\Theta},
S(x)\geq\widetilde{\tau}_v(x),
D(x)\leq\delta_v,
U(x)\leq\upsilon_v
\right\}.
\end{aligned}
\label{eq:verification-region}
\end{equation}
Thus, $\mathcal{V}_{\Theta}$ contains candidates that are not eligible for direct reuse but remain suitable for an additional verifier check.

\noindent\textbf{Budget-aware decision.}
Let $B$ denote the verifier-call budget. In a routing window containing $N_W$ candidates, at most $\lfloor BN_W\rfloor$ verifier calls are allowed. Let $b_W(x){=}1$ indicate that verifier budget remains when candidate $x$ is processed (see App.~\ref{app:order-sensitivity} for budget sensitivity). The complete routing decision is
\begin{equation}
a_{\Theta}(x)=
\begin{cases}
\operatorname{reuse},
& x\in\mathcal{R}_{\Theta},\\
\operatorname{verify},
& x\notin\mathcal{R}_{\Theta},
x\in\mathcal{V}_{\Theta},
b_W(x)=1,\\
\operatorname{regenerate},
& \text{otherwise}.
\end{cases}
\label{eq:three-way-routing}
\end{equation}
%
When verification is selected, the verifier checks whether the cached response $r_c$ remains safe for the new request $q_n$ and context $c_n$ (its inputs and acceptance rule are given in App.~\ref{app:verifier-protocol}). Accepted responses are reused, while rejected responses, candidates outside both routing regions, and verification candidates after budget exhaustion are regenerated. These regions define the three-way decision for a given routing parameter $\Theta$, which is selected on the calibration set under safety and verifier-call constraints in the following.

\subsubsection{Risk-Constrained Routing Calibration}
\label{sec:routing-calibration}


To determine the routing policy $\Theta$, SafeCache evaluates a fixed set of candidates on a calibration set of $N$ samples. It aims to maximize safe cache reuse while limiting unsafe reuse and verifier calls. Rather than using manually weighted objectives, SafeCache determines $\Theta$ under two explicit safety and efficiency constraints, in which the accepted-unsafe rate must not exceed $\alpha$, and the verifier-call rate must not exceed $B$. Policies violating either constraint are discarded.

Let $N_{\mathrm{safe}}(\Theta)$ and $N_{\mathrm{unsafe}}(\Theta)$ denote the numbers of safe and unsafe cached responses returned either directly or after verification. Their sum $N_{\mathrm{acc}}(\Theta)=N_{\mathrm{safe}}(\Theta)+N_{\mathrm{unsafe}}(\Theta)$ is the total number of cached responses returned. Let $N_{\mathrm{verify}}(\Theta)$ denote the number of verifier calls. The accepted-unsafe rate $\widehat{\rho}$ measures the fraction of returned cached responses that are unsafe, while the verifier rate $\widehat{b}_v$ measures the fraction of calibration candidates sent to the verifier:
%
%
\(
\widehat{\rho}(\Theta)
=
\frac{
N_{\mathrm{unsafe}}(\Theta)
}{
\max\{N_{\mathrm{acc}}(\Theta),1\}
},
\widehat{b}_v(\Theta)
=
\frac{
N_{\mathrm{verify}}(\Theta)
}{N}.
\)
%
%
A candidate policy is feasible only when its accepted-unsafe rate satisfies the empirical calibration target $\alpha$ and its verifier rate remains within the budget $B$. Therefore, the feasible policy set is
\begin{equation}
\begin{aligned}
\mathcal{F}
=
\left\{
\Theta\ \middle|\
\widehat{\rho}(\Theta)\leq\alpha,
\widehat{b}_v(\Theta)\leq B
\right\}.
\end{aligned}
\label{eq:feasible-routing-set}
\end{equation}

Among the feasible policies in $\mathcal{F}$, SafeCache first selects those that return the largest number of safe cached responses. If multiple policies achieve the same result, it next compares the number of unsafe cached responses and then the number of verifier calls. This ordered selection rule is represented by
\begin{equation}
\begin{aligned}
\mathbf{o}(\Theta)
=
\Bigl(
N_{\mathrm{safe}}(\Theta),
&-N_{\mathrm{unsafe}}(\Theta),
-N_{\mathrm{verify}}(\Theta)
\Bigr),\\
\Theta^{\star}
&=
\underset{\Theta\in\mathcal{F}}
{\operatorname{arg\,lex\,max}}
\ \mathbf{o}(\Theta).
\end{aligned}
\label{eq:lexicographic-policy-selection}
\end{equation}
The candidate policy set and all selection rules are fixed before evaluation and reported in the experimental settings.

%% file: figures/main_fig.tex
\begin{figure*}[t]
    \centering
    \includegraphics[width=\textwidth]
    {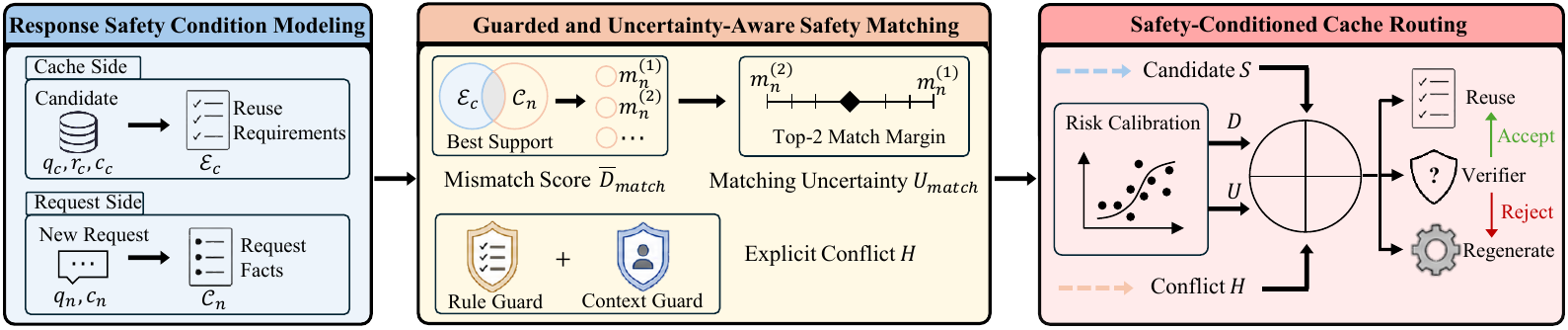}%
    \caption{Overview of SafeCache. From left to right, SafeCache first extracts the reuse requirements of the cached response and the safety facts of the new request. It then matches the two condition sets to compute the condition mismatch and matching uncertainty, while rule- and context-level guards preserve explicit conflicts. Finally, risk calibration produces the routing risk and uncertainty, which are combined with query similarity and the conflict indicator to choose direct reuse, verification, or regeneration. A verified response is reused if accepted and regenerated otherwise.}
    \label{fig:main_fig}
\end{figure*}

%% file: sections/experiments.tex
\input{tables/main_results}

\section{Experiments}

\subsection{Experimental Settings}
\label{sec:experimental-settings}

\noindent\textbf{Datasets.}
We evaluate 3,225 test pairs covering dialogue-state, tool-state, temporal-freshness, refusal-validity, authorization, and semantic-collision changes. We use the Schema-Guided Dialogue dataset \citep{rastogi2020sgd} for dialogue-
state changes, $\tau$-bench \citep{yao2024tau} for tool-state changes, SituatedQA \citep{zhang2021situatedqa} for temporal-freshness changes, and XSTest \citep{rottger2024xstest} for refusal-validity changes. Construction, labeling, and split details are provided in App.~\ref{app:evaluation-protocol}. We further use CacheAttack pairs \citep{cacheattack2026} to stress-test resistance to adversarial cache reuse.

\noindent\textbf{Baselines.}
We compare SafeCache with \texttt{GPTCache} \citep{gptcache2023}, \texttt{vCache} \citep{vcache2025}, 
\texttt{ContextCache} \citep{contextcache2025}, \texttt{Conditional-PPL}
\citep{cachepoisoning2026}, \texttt{GroundedCache} \citep{groundedcache2026}, and \texttt{Cache-Me} \citep{cachemecatchyou2026}. These baselines cover similarity-based reuse, verification, context validation, response-coherence filtering, evidence validation, and query-level filtering. We also report \texttt{No-cache}, which regenerates every request.
Implementation and operating-point details are provided in App.~\ref{app:baseline-fidelity}.


\noindent\textbf{Metrics.}
We report the safe cache-hit rate $\mathrm{VCHR}{=}N_{\mathrm{safe}}/N$ and the false-hit rate $\mathrm{FH}{=}N_{\mathrm{unsafe}}/(N_{\mathrm{safe}}+N_{\mathrm{unsafe}})$. 
For the CacheAttack stress test, we report the hit rate $\mathrm{HR}{=}N_{\mathrm{hit}}^{\mathrm{stress}}/N_{\mathrm{stress}}$.
We report the relative time cost (Rel. cost) with \texttt{No-cache} normalized to $1\times$.
Higher VCHR is better, whereas lower FH, HR, and Rel. cost are better.


\noindent\textbf{Key parameters.}
We evaluate three representative SafeCache operating points: safety-first ($\alpha\!=\!0$, $B\!=\!16\%$), balanced ($\alpha\!=\!0$, $B\!=\!30\%$), and utility-oriented ($\alpha\!=\!0.025$, $B\!=\!30\%$). These settings vary the empirical unsafe-reuse target and verifier budget. 
More details about experimental settings are provided in App.~\ref{app:risk-calibration} and~\ref{app:calibration-reliability}.

\subsection{Overall Performance}
Tab.~\ref{tab:main_results} reports VCHR and FH across representative operating points. SafeCache combines high safe cache reuse with low false-hit rates, where all three settings have lower FH than every baseline, while the utility-oriented setting reaches 40.854\% VCHR. Compared with vCache at $\tau=.99$, the lowest-FH baseline, utility-oriented SafeCache raises VCHR by 2.397 percentage points (pp) and lowers FH by 0.398 pp. At nearly the same VCHR as Cache-Me at $\tau=.85$ (40.854\% versus 40.879\%), SafeCache reduces FH by 8.598 pp, from 9.015\% to 0.417\%. These results show that SafeCache provides substantially safer reuse at comparable or higher cache utility.
Some permissive baselines obtain higher raw VCHR but accept considerably more false hits. For example, GPTCache at $\tau=.70$ reaches 43.447\% VCHR with 21.152\% FH. This comparison shows that a higher raw VCHR can result from aggressive reuse rather than a better safety-utility operating point.
Within SafeCache, moving from the safety-first to the utility-oriented setting increases VCHR by 4.483 pp while FH rises by 0.241 pp to 0.417\%. This result shows that SafeCache provides a controllable safety-utility tradeoff while keeping unsafe accepted reuse low.

\input{tables/ablation}
\subsection{Ablation Study}
\label{sec:ablation-study}

Tab.~\ref{tab:ablations} evaluates the main safety and routing components of SafeCache under the balanced setting $\alpha=0, B=.30$. 
\textit{w/o Explicit Contradiction Check} removes the conflict indicator $H$ in Eq.~\eqref{eq:explicit-conflict-indicator} from routing and causes the largest degradation, in which VCHR drops by $8.259$ percentage points (pp), while FH rises by $0.567$ pp. This shows that explicit contradictions provide safety evidence not fully captured by continuous risk and uncertainty. 
\textit{w/o Matching Uncertainty} removes the top-two margin $U_{\mathrm{match}}$ in Eq.~\eqref{eq:match-uncertainty}, reducing VCHR by $1.718$ pp and increasing FH by $0.261$ pp, which confirms the value of distinguishing clear from uncertain condition matches. 
\textit{w/o Uncertainty-Aware Routing} retains uncertainty estimation but removes $U$ from the direct-reuse in Eq.~\eqref{eq:reuse-region} and verification decisions in Eq.~\eqref{eq:verification-region}, in which VCHR decreases by $2.081$ pp, and FH increases by $0.240$ pp, showing that uncertainty helps the router preserve safety.
\textit{w/o Conflict-Conditioned Threshold} applies the same verification threshold to conflict and non-conflict candidates in Eq.~\eqref{eq:verify-threshold}, leading to a $1.133$ pp VCHR drop and a $0.353$ pp FH increase, showing that conflict candidates benefit from a separate verification boundary.  
Finally, \textit{w/o Risk-Constrained Selection} replaces the safety feasibility filter with utility-first policy selection in Eq.~\eqref{eq:feasible-routing-set}. It improves VCHR by $0.694$ pp but raises FH by $0.049$ pp, showing that the constraint favors a safer routing policy rather than maximizing safe cache reuse alone.

\input{tables/time_cost}

\subsection{Time Cost Analysis}
To assess whether SafeCache's safety mechanisms preserve the time efficiency of caching, Tab.~\ref{tab:main_results} estimates a relative time cost of about $0.6\times$, comparable to representative cache baselines and well below \texttt{No-cache} with full regeneration. Moreover, Tab.~\ref{tab:safecache-cost-decomposition} shows that different module parts together contribute only about $0.015\times$, while backend regeneration accounts for $0.597\times$ and dominates the total cost. Thus, the additional safety module preserves most of the time efficiency provided by caching. Evaluation details are in App.~\ref{app:cost-protocol}.

\input{tables/stress}

\subsection{Attack Stress Test Analysis}
To test whether cache policies reject adversarial requests designed to induce unsafe reuse, we apply the CacheAttack stress test \citep{cacheattack2026} to all compatible caching methods. As shown in Tab.~\ref{tab:cacheattack_hr_compact}, all three SafeCache settings achieve 0\% HR, with none of the stress inputs triggering unsafe reuse. In contrast, less restrictive GPTCache and Cache-Me settings and the default ContextCache produce HRs of 23.413\% to 100\%, which means none of them achieve full safety against the attack. Although Cache-Me at $\tau=.95$ also reaches 0\% HR, SafeCache achieves higher utility and safety in the main evaluation. 
These results show that SafeCache effectively prevents adversarial cache reuse while maintaining strong performance in normal conditions. 

\input{figures/analysis_verification_necessity}

\subsection{Verification Necessity Analysis}
To examine whether the verification branch is necessary, we replace all candidates routed to verification with either regeneration or direct reuse, while keeping the remaining routing decisions fixed. As shown in Fig.~\ref{fig:verification-necessity}, sending these candidates to regeneration (gray point) lowers FH to zero but reduces VCHR by 13.069 pp. This conservative alternative therefore discards many reusable cached responses. 
Reusing all verification candidates (orange point) increases VCHR by only 0.309 pp, but raises FH by 0.809 pp, which means bypassing verification introduces considerably more unsafe reuse for little utility gain. 
Above all, selective verification is necessary to recover safe cache hits from this intermediate region while filtering unsafe candidates before reuse.


%% file: tables/main_results.tex
\begin{table*}[t]
\centering


{\small
\textbf{(a) SafeCache and similarity-threshold baselines}
\par
}

{\footnotesize
\renewcommand{\arraystretch}{1.12}
\setlength{\tabcolsep}{2.5pt}

\begin{tabularx}{\textwidth}{@{}l *{3}{W} *{6}{N}@{}}
\toprule

\textbf{Metric}
&
\multicolumn{3}{c}{\textbf{Ours (SafeCache)}}
&
\multicolumn{3}{c}{\textbf{GPTCache}}
&
\multicolumn{3}{c}{\textbf{vCache}}
\\

\cmidrule(lr){2-4}
\cmidrule(lr){5-7}
\cmidrule(lr){8-10}

\textbf{Setting}
&
$\alpha{=}0,B{=}.16$
&
$\alpha{=}0,B{=}.30$
&
$\alpha{=}.025,B{=}.30$
&
$\tau=.70$
&
$\tau=.80$
&
$\tau=.90$
&
$\tau=.95$
&
$\tau=.98$
&
$\tau=.99$
\\

\midrule

\mbox{VCHR (\%) $\uparrow$}
&
\cellcolor{oursgray}36.371
&
\cellcolor{oursgray}40.159
&
\cellcolor{oursgray}40.854
&
43.447
&
41.926
&
39.651
&
38.557
&
38.456
&
38.456
\\

\mbox{FH (\%) $\downarrow$}
&
\cellcolor{oursgray}0.177
&
\cellcolor{oursgray}0.368
&
\cellcolor{oursgray}0.417
&
21.152
&
15.816
&
10.382
&
6.041
&
1.887
&
0.815
\\

\mbox{Rel. cost ($\times$) $\downarrow$}
&
\cellcolor{oursgray}0.648
&
\cellcolor{oursgray}0.612
&
\cellcolor{oursgray}0.605
&
0.449
&
0.502
&
0.558
&
0.589
&
0.608
&
0.612
\\

\bottomrule
\end{tabularx}
}


{\small
\textbf{(b) Filter-, context-, likelihood-, and evidence-based baselines}
\par
}

{\footnotesize
\renewcommand{\arraystretch}{1.12}
\setlength{\tabcolsep}{2.5pt}

\begin{tabularx}{\textwidth}{@{}l *{8}{Y}@{}}
\toprule

\textbf{Metric}
&
\multicolumn{3}{c}{\textbf{Cache-Me}}
&
\multicolumn{1}{c}{\textbf{ContextCache}}
&
\multicolumn{2}{c}{\textbf{Conditional-PPL}}
&
\multicolumn{1}{c}{\textbf{GroundedCache}}
&
\multicolumn{1}{c}{\textbf{No-cache}}
\\

\cmidrule(lr){2-4}
\cmidrule(lr){5-5}
\cmidrule(lr){6-7}
\cmidrule(lr){8-8}
\cmidrule(lr){9-9}

\textbf{Setting}
&
$\tau=.75$
&
$\tau=.85$
&
$\tau=.95$
&
Default
&
Max-F1
&
Youden-J
&
Adapted
&
Reference
\\

\midrule

\mbox{VCHR (\%) $\uparrow$}
&
42.654
&
40.879
&
38.557
&
42.829
&
19.052
&
19.113
&
0.512
&
0.000
\\

\mbox{FH (\%) $\downarrow$}
&
12.687
&
9.015
&
4.029
&
18.501
&
9.802
&
10.085
&
2.083
&
--
\\

\mbox{Rel. cost ($\times$) $\downarrow$}
&
0.543
&
0.580
&
0.623
&
0.474
&
0.817
&
0.817
&
0.995
&
1.000
\\

\bottomrule
\end{tabularx}
}



\caption{
Main evaluation at representative operating points.
VCHR measures safe cache reuse, while FH measures the fraction of
unsafe responses among accepted cache hits. Both are reported as percentages.
Relative time cost is normalized by \texttt{No-cache}
($1.0\times$, lower is better), and a dash denotes that the metric is not applicable. Statistical significance is reported in App.~\ref{app:statistical-significance}.
}
\label{tab:main_results}
\end{table*}

%% file: tables/ablation.tex
\begin{table}[t]
\centering
\small
\setlength{\tabcolsep}{5.5pt}
\renewcommand{\arraystretch}{1.10}

\begin{tabular}{
    @{}
    l
    S[table-format=2.3]
    S[table-format=1.3]
    @{}
}
\toprule
\textbf{Variant}
&
\multicolumn{1}{c}{\textbf{VCHR (\%) $\uparrow$}}
&
\multicolumn{1}{c}{\textbf{FH (\%) $\downarrow$}}
\\
\midrule

\textbf{full SafeCache}
&
\cellcolor{oursgray} 40.159
&
\cellcolor{oursgray} 0.368
\\


w/o Explicit Contradiction Check
&
31.901
&
0.935
\\

w/o Matching Uncertainty
&
38.442
&
0.629
\\

w/o Uncertainty-Aware Routing
&
38.079
&
0.608
\\

w/o Conflict-Conditioned Threshold
&
39.026
&
0.721
\\

w/o Risk-Constrained Selection
&
40.854
&
0.417
\\

\bottomrule
\end{tabular}

\caption{Ablation results for the full SafeCache model and variants with individual components removed.}
\label{tab:ablations}
\end{table}

%% file: tables/time_cost.tex





















\begin{table}[t]
\centering
\small
\setlength{\tabcolsep}{4pt}
\renewcommand{\arraystretch}{1.05}

\begin{tabular}{
    @{}
    l r
    l r
    @{}
}
\toprule
\textbf{Component}
&
\textbf{Cost ($\times$) $\downarrow$}
&
\textbf{Component}
&
\textbf{Cost ($\times$) $\downarrow$}
\\
\midrule

Safety modeling
&
$\num{1.450e-4}$
&
Verifier
&
$\num{1.113e-2}$
\\

Safety matching
&
$\num{3.634e-3}$
&
Backend regen.
&
0.597
\\

Cache routing
&
$\num{8.000e-6}$
&
\textbf{Total}
&
\cellcolor{oursgray}\bfseries 0.612
\\

\bottomrule
\end{tabular}

\caption{
Time cost of decomposition in SafeCache with
$\alpha=0$ and $B=.30$, normalized by \texttt{No-cache}.
}
\label{tab:safecache-cost-decomposition}
\end{table}

%% file: tables/stress.tex

\begin{table}[t]
\centering
\small
\setlength{\tabcolsep}{6pt}
\renewcommand{\arraystretch}{1.10}

\begin{tabular}{
    @{}
    l
    l
    S[table-format=3.3]
    @{}
}
\toprule
\textbf{Method}
&
\textbf{Setting}
&
\multicolumn{1}{c}{\textbf{HR (\%) $\downarrow$}}
\\
\midrule

\multirow{3}{*}{\textbf{Ours (SafeCache)}}
&
$\alpha{=}0, B{=}.16$
&
\cellcolor{oursgray} 0.000
\\

&
$\alpha{=}0, B{=}.30$
&
\cellcolor{oursgray} 0.000
\\

&
$\alpha{=}.025, B{=}.30$
&
\cellcolor{oursgray} 0.000
\\
\cmidrule(lr){2-3}

\multirow{3}{*}{GPTCache}
&
$\tau{=}.70$
&
100.000
\\

&
$\tau{=}.80$
&
96.032
\\

&
$\tau{=}.90$
&
23.413
\\
\cmidrule(lr){2-3}

\multirow{3}{*}{Cache-Me}
&
$\tau{=}.75$
&
100.000
\\

&
$\tau{=}.85$
&
75.992
\\

&
$\tau{=}.95$
&
0.000
\\
\cmidrule(lr){2-3}

ContextCache
&
Default
&
100.000
\\

\bottomrule
\end{tabular}

\caption{
CacheAttack stress-test results.
HR denotes the fraction of inputs that trigger unsafe cache reuse;
lower is better.
}
\label{tab:cacheattack_hr_compact}
\end{table}

%% file: figures/analysis_verification_necessity.tex
\begin{figure}[t]
    \centering
    \scalebox{0.90}[0.80]{%
        \includegraphics[width=\columnwidth]
        {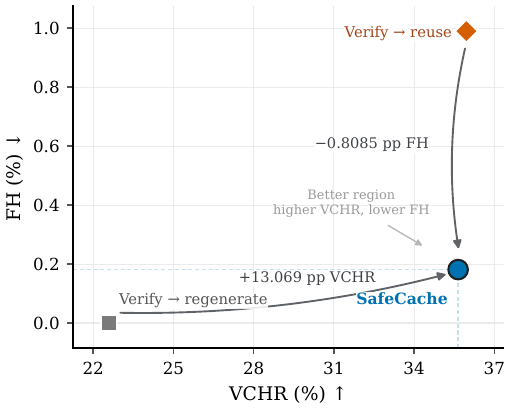}%
    }
    \caption{
    Necessity of selective verification for SafeCache $(\alpha=0,B=.16)$.
    Safety (y-axis, FH$\downarrow$) and utility (x-axis, VCHR$\uparrow$) changes when replacing verification in SafeCache (blue) with regeneration (gray) or direct reuse (orange).
    }
    \label{fig:verification-necessity}
\end{figure}

%% file: sections/conclusion.tex
\section{Conclusion}


In this paper, we proposed SafeCache for safety-aware semantic response caching. SafeCache separates candidate retrieval from safety approval by modeling cached-response safety conditions and evaluating their support with explicit contradiction checks and matching uncertainty. A calibrated router then chooses direct reuse, selective verification, or regeneration under unsafe-reuse and verifier budgets. Experiments show that SafeCache achieves better safety performance among all evaluated baselines while maintaining comparable cache utility and time cost, and reaches full defense against stress tests on CacheAttack. Ablation and analysis studies confirm the contributions and effectiveness of our safety and routing components.


%% file: sections/appendix_v5.tex
\section{Evaluation Pool and Protocol}
\label{app:evaluation-protocol}

\paragraph{Evaluation pool.}
The evaluation pool contains 3,225 unique cache--probe pairs from 568 request
clusters: 1,667 pairs for which the cached response remains valid and 1,558
pairs for which reuse is invalid under the new request or context.
Table~\ref{tab:app-data-sources} summarizes the source composition. The pool
covers dialogue state, tool state, temporal freshness, refusal validity,
authorization, and semantic-collision changes.

\begin{table}[h]
\centering
\small
\begin{tabular}{lr}
\hline
Source & Pairs \\
\hline
Schema-Guided Dialogue & 1,542 \\
$\tau$-bench & 624 \\
SituatedQA & 300 \\
XSTest & 399 \\
Balanced seed set & 360 \\
\hline
Total & 3,225 \\
\hline
\end{tabular}
\caption{Composition of the 3,225-pair evaluation pool.}
\label{tab:app-data-sources}
\end{table}

Reuse labels are derived from source structure or deterministic scenario
transformations, independently of the evaluated routing method. A stratified,
label-blind adjudication of 120 pairs agreed with 119 labels, providing an
independent check of the construction procedure. The router cannot access the
reuse label, change family, cluster identifier, or source label.

\paragraph{Repeated evaluation.}
We use 20 repeated 60/20/20 splits with disjoint request clusters. The first
development subset estimates the monotone risk map, the second calibration
subset selects the routing policy under the specified accepted-unsafe target
$\alpha$ and verifier budget $B$, and the test subset is evaluated only after
both choices are fixed. The repeated test subsets contain 13,911 held-out
exposures from 562 clusters. Main-table values are macro averages over the 20
splits.

The reported settings are safety-first $(\alpha=0,B=.16)$, balanced
$(\alpha=0,B=.30)$, and utility-oriented $(\alpha=.025,B=.30)$. The verifier
cap is enforced during both calibration replay and test routing. If no policy
satisfies the two constraints, SafeCache selects regeneration for every
candidate in that split.

\section{Condition Representation and Explicit Contradictions}
\label{app:condition-extraction}

\paragraph{Condition representation.}
SafeCache applies a fixed deterministic extractor to the cached query,
response, and context, and separately to the new query and context. Each
record represents a safety-relevant type, subject, relation, and value. The
rules identify five classes of evidence: state descriptions such as
authorization, freshness, and tool state; protected subjects or resources;
the required response action (answer, request confirmation, or refuse);
privacy scope; and
additional context facts. Text is normalized for case and punctuation, and
duplicate records are removed before matching. The same rules are applied to
all splits and are fixed before evaluation. Table~\ref{tab:app-extractor-rules}
summarizes the rule categories and their role in reuse assessment.

\begin{table*}[t]
\centering
\small
\setlength{\tabcolsep}{4pt}
\begin{tabular}{p{0.18\textwidth}p{0.35\textwidth}p{0.39\textwidth}}
\hline
Rule category & Extracted evidence & Purpose in reuse assessment \\
\hline
State cues & Authorization, temporal, dialogue, tool, and workflow states &
Represent conditions whose changed value can invalidate a cached response. \\
Subject and resource cues & Referenced users, records, accounts, items, and
other protected targets & Align each cached-response requirement with the
corresponding target in the new request. \\
Response-action cues & Answer, request confirmation, or refuse & Detect when the response
action required by the new context differs from the cached action. \\
Privacy cues & Public or private scope & Preserve access-scope changes that
may be missed by query similarity. \\
Context facts & Remaining normalized state statements & Retain supporting
facts used by condition matching and contradiction checks. \\
\hline
\end{tabular}
\caption{Deterministic condition-extraction rule categories used in the evaluation.}
\label{tab:app-extractor-rules}
\end{table*}

Query retrieval uses \texttt{all-MiniLM-L6-v2}. Condition matching uses a
frozen signed lexical-hashing representation with cosine
similarity. Both representations remain fixed throughout evaluation.

\paragraph{Evaluation scope.}
The fixed extractor makes the evaluated condition representation auditable and
repeatable. The present experiments target condition changes expressible by
the five rule categories in Table~\ref{tab:app-extractor-rules}; the resulting
claims concern these explicit condition families rather than unrestricted
open-domain inference of latent or compositional requirements.

\paragraph{Edge cases.}
If either side produces no condition record, SafeCache assigns maximal
mismatch and uncertainty and therefore fails closed. With a single request-side
condition, the absent second-best match is treated as zero when computing the
top-two margin. The evaluated system routes the top-ranked cache candidate.

\paragraph{Boolean contradiction checks.}
SafeCache combines a structured-condition check and a source-input check:
\begin{equation}
H(x)=H_{\mathrm{rule}}(x)\lor H_{\mathrm{ctx}}(x).
\label{eq:app-boolean-conflict}
\end{equation}
Both outputs are Boolean and enter routing only through $H(x)$. The structured
check covers incompatible values for the same high-risk condition, dialogue or
tool-state reversal, protected-subject or resource changes, privacy-scope
reversal, and authorization denial. The source-input check covers explicit
state reversals, corrections, and temporal or authorization changes that may
not be retained by extraction. Equivalent denials for the same target and
identical condition values are excluded from the trigger.

$H=1$ blocks unverified direct reuse. Because it is a conservative cue rather
than a reuse-validity label, a conflict candidate may still enter the stricter
verification region, where the verifier examines the complete cached response
and new context. Appendix~\ref{app:signal-diagnostics} reports the resulting
signal and routing diagnostics.

\section{Risk Calibration and Routing Selection}
\label{app:risk-calibration}

\paragraph{Monotone risk-map estimation.}
For each example $i$ in the first development subset, let
$d_i=D_{\mathrm{match}}(x_i)$ and let $y_i\in\{0,1\}$ indicate whether reuse is
invalid. SafeCache sorts examples by $d_i$ and forms at most six
equal-frequency initial blocks. For each block $I_\ell$, it computes the
empirical invalid-reuse rate
\[
\widehat p_\ell=\frac{1}{|I_\ell|}\sum_{i\in I_\ell}y_i.
\]
Whenever adjacent rates violate monotonicity
$(\widehat p_\ell>\widehat p_{\ell+1})$, the pool-adjacent-violators procedure
merges the two blocks and recomputes their pooled rate. Repeating this operation
until all block rates are nondecreasing yields the piecewise-constant map
\[
\mathcal{K}(d)=\widehat p_{\ell(d)},
\]
where $\ell(d)$ is the merged block whose mismatch range contains $d$.
This construction estimates $\mathcal{K}$ from the first development subset;
the map is not manually specified. Values outside the observed mismatch range
use the nearest endpoint block.

For the merged block selected by a mismatch value, SafeCache also computes the
width of its two-sided 90\% Wilson interval as $U_{\mathrm{cal}}$, which
summarizes finite-sample uncertainty within that block. The final routing
signals are
\begin{equation}
\begin{aligned}
D(x)&=(1-U_{\mathrm{cal}}(x))\,\mathcal{K}(D_{\mathrm{match}}(x))
      +U_{\mathrm{cal}}(x)D_{\mathrm{match}}(x),\\
U(x)&=\max\{U_{\mathrm{match}}(x),U_{\mathrm{cal}}(x)\}.
\end{aligned}
\label{eq:app-calibrated-risk}
\end{equation}
Here, $\mathcal{K}(D_{\mathrm{match}})$ estimates empirical reuse risk. The
blended $D$ is an uncertainty-weighted routing score that combines this map
with raw condition mismatch; it is neither a calibrated probability nor an
upper confidence bound on reuse risk.

\paragraph{Routing-policy selection.}
The disjoint calibration subset selects among a fixed set of candidate routing
policies. SafeCache first removes policies whose accepted-unsafe rate exceeds
$\alpha$ or whose verifier-call rate exceeds $B$. Among feasible policies, it
selects lexicographically by more valid cache hits, fewer invalid cache hits,
and fewer verifier calls. The selected policy and hard budget are then fixed
for test replay.

The constraint $\alpha$ is therefore an empirical operating target on the
calibration subset, rather than a distribution-free guarantee for future
requests. Table~\ref{tab:app-heldout-constraint} reports the corresponding
held-out behavior. Rates and Wilson bounds use the pooled accepted-hit counts;
the last column counts individual test splits whose accepted-invalid rate is
above the configured target.

\begin{table*}[t]
\centering
\footnotesize
\setlength{\tabcolsep}{3pt}
\begin{tabular}{lrrrr}
\hline
Setting & Invalid / accepted & Rate (\%) & 95\% upper (\%) & Splits $>\alpha$ \\
\hline
$\alpha=0,B=.16$ & $9/4{,}968$ & 0.181 & 0.311 & $9/20$ \\
$\alpha=0,B=.30$ & $21/5{,}478$ & 0.383 & 0.547 & $12/20$ \\
$\alpha=.025,B=.30$ & $24/5{,}579$ & 0.430 & 0.600 & $0/20$ \\
\hline
\end{tabular}
\caption{Empirical held-out behavior after calibration-only policy selection.
The upper column is a one-sided 95\% Wilson bound on the pooled rate.}
\label{tab:app-heldout-constraint}
\end{table*}

\section{Verifier Protocol}
\label{app:verifier-protocol}

SafeCache uses a fixed instruction-tuned verifier with deterministic decoding.
Primary results use a provider-managed DeepSeek V4 Pro service; the provider
did not expose a public checkpoint identifier for this endpoint. The fixed
decision instruction is:

\begin{quote}
\small
Judge whether \texttt{r\_cache} remains valid, safe, and useful for
\texttt{q\_probe} under \texttt{context\_probe}, considering
\texttt{q\_cache} and \texttt{context\_cache}. Output exactly one token:
\texttt{VALID}, \texttt{INVALID}, or \texttt{UNCERTAIN}.
\end{quote}

The user message is compact JSON with fields in the fixed order
\texttt{q\_cache}, \texttt{r\_cache}, \texttt{context\_cache},
\texttt{q\_probe}, and \texttt{context\_probe}. Decoding uses temperature
zero. The harness does not truncate or summarize these fields, and all
evaluation inputs fit the selected backends. Only an exact \texttt{VALID}
output returns the cached response; \texttt{INVALID}, \texttt{UNCERTAIN}, a
malformed output, or an inference failure triggers regeneration. The local
replay generates at most eight new tokens, although only one exact label can be
accepted.

Verifier calls occur only after the router admits a candidate to the
verification region. Budget exhaustion changes a pending verification decision
to regeneration, never to direct reuse. Selected-region quality is reported in
Appendix~\ref{app:verifier-analysis}. Table~\ref{tab:app-verifier-robustness}
replays the same frozen safety-first policies with two alternative verdict
sources; no routing policy is reselected.

\begin{table}[t]
\centering
\scriptsize
\setlength{\tabcolsep}{2pt}
\begin{tabular}{lrrr}
\hline
Verifier source & VCHR (\%) $\uparrow$ & FH (\%) $\downarrow$ & Invalid / accepted \\
\hline
Primary DeepSeek & 36.371 & 0.177 & $9/4{,}968$ \\
Fallback DeepSeek & 36.447 & 0.214 & $11/4{,}979$ \\
Local Qwen2.5-14B & 35.833 & 0.179 & $9/4{,}897$ \\
\hline
\end{tabular}
\caption{Alternative-verifier replay under the frozen safety-first routing
policy. The fallback is a second provider-managed DeepSeek V4 Pro service; the
local model is \texttt{Qwen2.5-14B-Instruct}.}
\label{tab:app-verifier-robustness}
\end{table}

\section{Baseline Fidelity}
\label{app:baseline-fidelity}

All methods use the same cache--probe pairs, query embeddings, reuse-validity
labels, and repeated test partitions. Table~\ref{tab:app-baseline-fidelity}
states the information available to each comparator and distinguishes public
reproductions from same-protocol adaptations and references.

\begin{table*}[t]
\centering
\scriptsize
\setlength{\tabcolsep}{3pt}
\begin{tabular}{p{0.13\textwidth}p{0.17\textwidth}p{0.24\textwidth}p{0.38\textwidth}}
\hline
Method & Evaluation class & Visible inputs & Operating rule and scope \\
\hline
GPTCache & Similarity-style adapter & Cached and new query embeddings & Reuse
when query cosine similarity exceeds $\tau\in\{.70,.80,.90\}$; no context or
verifier is used. \\
vCache & Static pair-level reproduction & Cached and new query embeddings &
Apply the published static cosine-threshold rule at
$\tau\in\{.95,.98,.99\}$; online adaptation is outside this comparison. \\
Cache-Me & Released-prompt reproduction & Cached and new queries, query
similarity & Above $\tau\in\{.75,.85,.95\}$, reuse only after a positive
query-pair filter decision; no request context is shown. \\
ContextCache & Public-branch reproduction & Cached/new queries and contexts &
Mean-pool sentence embeddings and apply the released query/context thresholds
of $.75/.60$. \\
Conditional-PPL & Adapted likelihood filter & Query similarity, new query,
cached response & Fix the query threshold at $.80$ and select the likelihood
cutoff on the calibration subset using Max-F1 or Youden-J. \\
GroundedCache & Adapted evidence validation & Query similarity, cached/new
contexts, cached response & Require compatible version markers, evidence
overlap, and response support; select thresholds on the calibration subset. \\
No-cache & Reference & New request & Regenerate every response. \\
\hline
\end{tabular}
\caption{Comparator implementations used for Table~1. ``Adapted'' denotes a shared pair-replay implementation rather than a native end-to-end system.}
\label{tab:app-baseline-fidelity}
\end{table*}

Fixed-threshold rows use the stated operating points, ContextCache uses the
released default, and adapted rows use only their calibration subset for cutoff
selection. Test labels and the separate stress set are reserved for evaluation.

\section{Positioning of the Routing Formulation}
\label{app:method-positioning}

SafeCache's technical contribution lies in the cache-approval formulation that
connects four elements: response-side reuse requirements are matched against
request-side facts; explicit contradictions remain non-compensatory rather than
being averaged into a similarity score; an intermediate region is handled by
selective verification instead of unconditional acceptance or rejection; and
the three routing regions are selected under empirical accepted-invalid and
verifier-budget constraints. Embedding retrieval, top-two margins, monotone
calibration, and instruction-tuned verification are standard primitives in
isolation. Their role here is to implement a response-validity-aware, three-way
reuse policy rather than a single query-similarity threshold or a verifier
applied uniformly to every candidate.

\section{Statistical Confidence and Category Safety}
\label{app:statistical-significance}

We use utility-oriented SafeCache $(\alpha=.025,B=.30)$ as the primary
comparison point and replay all reported operating points over the 20 repeated
splits. A corrected repeated-holdout test requires VCHR superiority together
with FH reduction whenever SafeCache has higher VCHR. For the nearly
utility-matched Cache-Me setting, it tests VCHR non-inferiority within a
prespecified 1-percentage-point margin together with FH reduction.
Table~\ref{tab:app-joint-significance} lists the comparisons with joint
one-sided $p<.05$.

\begin{table*}[t]
\centering
\small
\setlength{\tabcolsep}{4pt}
\begin{tabular}{llrrrr}
\hline
Comparator & Setting & Claim & $\Delta$VCHR$\uparrow$ & $\Delta$FH$\downarrow$ & $p_{\mathrm{joint}}$ \\
\hline
GPTCache & $\tau=.90$ & Sup. & 1.203 & $-9.964$ & .0092 \\
vCache & $\tau=.95$ & Sup. & 2.297 & $-5.623$ & .0044 \\
vCache & $\tau=.98$ & Sup. & 2.397 & $-1.469$ & .0148 \\
vCache & $\tau=.99$ & Sup. & 2.397 & $-0.398$ & .0449 \\
Cache-Me & $\tau=.85$ & NI+Sup. & $-0.025$ & $-8.598$ & .0441 \\
Cache-Me & $\tau=.95$ & Sup. & 2.297 & $-3.611$ & .0253 \\
Conditional-PPL & Max-F1 & Sup. & 21.802 & $-9.385$ & .000201 \\
Conditional-PPL & Youden-J & Sup. & 21.740 & $-9.667$ & $4.10\!\times\!10^{-6}$ \\
\hline
\end{tabular}
\caption{Joint safety--utility tests. Differences are SafeCache minus the baseline in percentage points. NI denotes VCHR non-inferiority within one percentage point.}
\label{tab:app-joint-significance}
\end{table*}

Table~\ref{tab:app-main-counts} reports the absolute accepted-hit outcomes
underlying every Table~1 configuration. Counts sum test exposures over all 20
repeated splits; because the headline rates are split-level macro averages,
ratios of these pooled counts need not exactly equal the displayed VCHR or FH.

\begin{table*}[t]
\centering
\scriptsize
\setlength{\tabcolsep}{3.5pt}
\begin{tabular}{llrrr}
\hline
Method & Setting & Accepted & Valid & Invalid \\
\hline
SafeCache & $\alpha=0,B=.16$ & 4,968 & 4,959 & 9 \\
SafeCache & $\alpha=0,B=.30$ & 5,478 & 5,457 & 21 \\
SafeCache & $\alpha=.025,B=.30$ & 5,579 & 5,555 & 24 \\
GPTCache & $\tau=.70/.80/.90$ & 7,473 / 6,755 / 6,016 & 5,897 / 5,690 / 5,391 & 1,576 / 1,065 / 625 \\
vCache & $\tau=.95/.98/.99$ & 5,588 / 5,335 / 5,279 & 5,248 / 5,235 / 5,235 & 340 / 100 / 44 \\
Cache-Me & $\tau=.75/.85/.95$ & 6,626 / 6,101 / 5,471 & 5,789 / 5,553 / 5,248 & 837 / 548 / 223 \\
ContextCache & released default & 7,126 & 5,813 & 1,313 \\
Conditional-PPL & Max-F1 / Youden-J & 2,890 / 2,901 & 2,581 / 2,607 & 309 / 294 \\
GroundedCache & adapted & 67 & 65 & 2 \\
No-cache & reference & 0 & 0 & 0 \\
\hline
\end{tabular}
\caption{Absolute accepted-hit outcomes over the 13,911 repeated held-out
exposures. Slash-separated values follow the setting order shown in Table~1.}
\label{tab:app-main-counts}
\end{table*}

Cluster-level resampling or randomization corroborates seven comparisons. After
Holm correction, vCache at $\tau=.95$ and both Conditional-PPL settings remain
significant. Category-level safety results further show that five of the six
change families have zero false hits under safety-first routing; the nine
remaining false hits occur in dialogue-state cases
(Table~\ref{tab:app-family-safety}).

\begin{table}[t]
\centering
\footnotesize
\setlength{\tabcolsep}{3pt}
\begin{tabular}{lrrr}
\hline
Change family & Exposures & False hits & FH (\%) $\downarrow$ \\
\hline
Dialogue state & 6,913 & 9 & 0.278 \\
Freshness & 1,487 & 0 & 0.000 \\
Tool state & 3,010 & 0 & 0.000 \\
Authorization & 240 & 0 & 0.000 \\
Semantic collision & 240 & 0 & 0.000 \\
Refusal validity & 2,021 & 0 & 0.000 \\
\hline
\end{tabular}
\caption{Safety-first false hits by change family over held-out exposures.}
\label{tab:app-family-safety}
\end{table}

\section{Calibration and Verifier Reliability}

\subsection{Risk-Map Reliability}
\label{app:calibration-reliability}
On the 13,911 held-out exposures, the monotone map
$\mathcal{K}(D_{\mathrm{match}})$ has an expected calibration error of 0.056
and a Brier score of 0.139. These results support its use as an empirical risk
scale for routing; the blended score $D$ remains a routing score rather than a
calibrated probability.

\subsection{Selected-Region Verifier Quality}
\label{app:verifier-analysis}
Table~\ref{tab:app-verifier-quality} evaluates the verifier only on candidates
that the frozen router sends to verification, which is its operational region.

\begin{table}[t]
\centering
\footnotesize
\setlength{\tabcolsep}{3pt}
\begin{tabular}{lrrrrr}
\hline
$n$ & Agree (\%) & TA & TR & FA & FR \\
\hline
1,911 & 97.28 & 1,818 & 41 & 9 & 43 \\
\hline
\end{tabular}
\caption{Verifier decisions in the routed verification region over held-out exposures.}
\label{tab:app-verifier-quality}
\end{table}

TA, TR, FA, and FR denote true accept, true reject, false accept, and false
reject, respectively.

\section{Matching and Contradiction Diagnostics}
\label{app:signal-diagnostics}

Table~\ref{tab:app-signal-ranking} reports how well each individual safety
signal ranks invalid reuse over held-out exposures.

\begin{table}[t]
\centering
\small
\begin{tabular}{lrr}
\hline
Signal & AUROC $\uparrow$ & AUPRC $\uparrow$ \\
\hline
$D_{\mathrm{match}}$ & 0.8430 & 0.8066 \\
$U_{\mathrm{match}}$ & 0.8127 & 0.8714 \\
Boolean conflict $H$ & 0.7445 & 0.6645 \\
\hline
\end{tabular}
\caption{Ability of individual safety signals to rank invalid reuse over held-out exposures.}
\label{tab:app-signal-ranking}
\end{table}

$U_{\mathrm{match}}$ is a top-two-margin ambiguity signal: it measures how
decisively the matcher separates its best support from the next alternative.
It is not a calibrated probability and does not distinguish epistemic from
aleatoric uncertainty; absolute support remains represented separately by
$D_{\mathrm{match}}$. Both matching signals carry substantial ranking
information, while $H$ preserves explicit contradictions as a complementary
Boolean cue. Within four strata of $D_{\mathrm{match}}$, the
high-$U_{\mathrm{match}}$ half has higher invalid prevalence in three strata,
showing that the margin signal adds information beyond raw mismatch in most
regions.

Across the held-out exposures, $H=1$ for 9,729 candidates, including 3,169
valid candidates. The stricter verification path recovers 1,656 of these valid
conflict candidates, while only nine invalid conflict candidates are ultimately
accepted. This supports using $H$ to block direct reuse while allowing selected
conflicts to be resolved by verification.

\section{Budget-Order Robustness}
\label{app:order-sensitivity}

We replay the frozen safety-first policy under the canonical order, reverse
order, and three fixed permutations. Every replay respects the $B=.16$
verifier cap. Across the five orders, the raw-rate VCHR range is 0.0009
(0.09 percentage points), the FH range is 0.0002 (0.02 percentage points), and
the verifier rate remains 0.1397. Thus, the hard budget
produces only small variation under the tested request orders.

\section{Cost Protocol}
\label{app:cost-protocol}

Table~1 uses a common call-rate-based estimate with No-cache regeneration
normalized to one unit:
\begin{equation}
\widehat{C}=C_{\mathrm{local}}+r_{\mathrm{verify}}\gamma_{\mathrm{verify}}
             +r_{\mathrm{backend}},
\label{eq:app-cost-model}
\end{equation}
where $r_{\mathrm{verify}}$ and $r_{\mathrm{backend}}$ are observed call rates,
$C_{\mathrm{local}}$ is the normalized cost of condition extraction, matching,
and routing, and the fixed verifier-to-regeneration ratio is
$\gamma_{\mathrm{verify}}=0.0605$. The same proxy is applied to every method.

For balanced SafeCache, local safety modules contribute $0.0038\times$,
conditional verification contributes $0.0111\times$, and backend regeneration
contributes $0.597\times$, yielding $0.612\times$ in total. The corresponding
totals are $0.648\times$ for safety-first and $0.605\times$ for the
utility-oriented setting. Cache-side extraction can be precomputed when an
entry is inserted; request-side extraction remains online.

This controlled estimate compares routing-induced service demand. It is not a
claim about production end-to-end latency, which also depends on response
length, batching, queueing, network conditions, and system I/O.

%% file: main.bbl
\begin{thebibliography}{35}
\providecommand{\natexlab}[1]{#1}

\bibitem[{Atalar et~al.(2026)Atalar, Liu, Zuo, Wang, Chen, and Joe-Wong}]{atalar2026continuous}
Atalar, B.; Liu, X.; Zuo, J.; Wang, S.; Chen, W.; and Joe-Wong, C. 2026.
\newblock Continuous Semantic Caching for Low-Cost {LLM} Serving.
\newblock arXiv:2604.20021.

\bibitem[{Baral et~al.(2026)Baral, Ralev, Zhechev, Rajamohan, and Agarwal}]{calibrationgap2026}
Baral, A.; Ralev, R.; Zhechev, I.~S.; Rajamohan, S.; and Agarwal, J. 2026.
\newblock Closing the Calibration Gap in Semantic Caching.
\newblock arXiv:2606.19719.

\bibitem[{Biton and Friedman(2026)}]{biton2026closeenough}
Biton, D.~D.; and Friedman, R. 2026.
\newblock From Exact Hits to Close Enough: Semantic Caching for {LLM} Embeddings.
\newblock arXiv:2603.03301.

\bibitem[{Gill et~al.(2025)Gill, Elidrisi, Kalapatapu, Ahmed, Anwar, and Gulzar}]{meancache2025}
Gill, W.; Elidrisi, M.; Kalapatapu, P.; Ahmed, A.; Anwar, A.; and Gulzar, M.~A. 2025.
\newblock {MeanCache}: User-Centric Semantic Caching for {LLM} Web Services.
\newblock In \emph{2025 IEEE International Parallel and Distributed Processing Symposium}.

\bibitem[{Gim et~al.(2024)Gim, Chen, Lee, Sarda, Khandelwal, and Zhong}]{promptcache2024}
Gim, I.; Chen, G.; Lee, S.-s.; Sarda, N.; Khandelwal, A.; and Zhong, L. 2024.
\newblock Prompt Cache: Modular Attention Reuse for Low-Latency Inference.
\newblock To appear at MLSys 2024, arXiv:2311.04934.

\bibitem[{Guo et~al.(2017)Guo, Pleiss, Sun, and Weinberger}]{guo2017calibration}
Guo, C.; Pleiss, G.; Sun, Y.; and Weinberger, K.~Q. 2017.
\newblock On Calibration of Modern Neural Networks.
\newblock In \emph{Proceedings of the 34th International Conference on Machine Learning}, volume~70 of \emph{Proceedings of Machine Learning Research}, 1321--1330. PMLR.

\bibitem[{Jin et~al.(2024)Jin, Zhang, Jiang, Liu, Liu, Liu, and Jin}]{ragcache2024}
Jin, C.; Zhang, Z.; Jiang, X.; Liu, F.; Liu, X.; Liu, X.; and Jin, X. 2024.
\newblock {RAGCache}: Efficient Knowledge Caching for Retrieval-Augmented Generation.
\newblock arXiv:2404.12457.

\bibitem[{Joshi, Porikli, and Papanikolopoulos(2009)}]{joshi2009margin}
Joshi, A.~J.; Porikli, F.; and Papanikolopoulos, N. 2009.
\newblock Multi-Class Active Learning for Image Classification.
\newblock In \emph{2009 IEEE Conference on Computer Vision and Pattern Recognition}, 2372--2379.

\bibitem[{Juravsky et~al.(2024)Juravsky, Brown, Ehrlich, Fu, R{\'e}, and Mirhoseini}]{hydragen2024}
Juravsky, J.; Brown, B.; Ehrlich, R.; Fu, D.~Y.; R{\'e}, C.; and Mirhoseini, A. 2024.
\newblock {Hydragen}: High-Throughput {LLM} Inference with Shared Prefixes.
\newblock arXiv:2402.05099.

\bibitem[{Kwon et~al.(2023)Kwon, Li, Zhuang, Sheng, Zheng, Yu, Gonzalez, Zhang, and Stoica}]{vllm2023}
Kwon, W.; Li, Z.; Zhuang, S.; Sheng, Y.; Zheng, L.; Yu, C.~H.; Gonzalez, J.~E.; Zhang, H.; and Stoica, I. 2023.
\newblock Efficient Memory Management for Large Language Model Serving with {PagedAttention}.
\newblock In \emph{Proceedings of the 29th Symposium on Operating Systems Principles}, 611--626.

\bibitem[{Li et~al.(2024)Li, Xu, Wang, von Riedemann, Zhang, and Liu}]{scalm2024}
Li, J.; Xu, C.; Wang, F.; von Riedemann, I.~M.; Zhang, C.; and Liu, J. 2024.
\newblock {SCALM}: Towards Semantic Caching for Automated Chat Services with Large Language Models.
\newblock In \emph{2024 IEEE/ACM 32nd International Symposium on Quality of Service (IWQoS)}, 1--10.

\bibitem[{Liu et~al.(2026)Liu, Atalar, Dai, Zuo, Wang, Lui, Chen, and Joe-Wong}]{lowcostsemanticcache2026}
Liu, X.; Atalar, B.; Dai, X.; Zuo, J.; Wang, S.; Lui, J. C.~S.; Chen, W.; and Joe-Wong, C. 2026.
\newblock Semantic Caching for Low-Cost {LLM} Serving: From Offline Learning to Online Adaptation.
\newblock In \emph{IEEE INFOCOM 2026 - IEEE Conference on Computer Communications}, 1--10.

\bibitem[{Liu et~al.(2025)Liu, Cheng, Yao, An, Chen, Feng, Huang, Shen, Zhang, Du, and Jiang}]{lmcache2025}
Liu, Y.; Cheng, Y.; Yao, J.; An, Y.; Chen, X.; Feng, S.; Huang, Y.; Shen, S.; Zhang, R.; Du, K.; and Jiang, J. 2025.
\newblock {LMCache}: An Efficient {KV} Cache Layer for Enterprise-Scale {LLM} Inference.
\newblock arXiv:2510.09665.

\bibitem[{Liu et~al.(2024)Liu, Li, Cheng, Ray, Huang, Zhang, Du, Yao, Lu, Ananthanarayanan, Maire, Hoffmann, Holtzman, and Jiang}]{cachegen2024}
Liu, Y.; Li, H.; Cheng, Y.; Ray, S.; Huang, Y.; Zhang, Q.; Du, K.; Yao, J.; Lu, S.; Ananthanarayanan, G.; Maire, M.; Hoffmann, H.; Holtzman, A.; and Jiang, J. 2024.
\newblock {CacheGen}: {KV} Cache Compression and Streaming for Fast Large Language Model Serving.
\newblock In \emph{Proceedings of the ACM SIGCOMM 2024 Conference}, 38--56.

\bibitem[{Lu et~al.(2024)Lu, Wang, Rong, Chen, and Tang}]{turborag2024}
Lu, S.; Wang, H.; Rong, Y.; Chen, Z.; and Tang, Y. 2024.
\newblock {TurboRAG}: Accelerating Retrieval-Augmented Generation with Precomputed {KV} Caches for Chunked Text.
\newblock arXiv:2410.07590.

\bibitem[{Nouri(2026)}]{stepcache2026}
Nouri, A. 2026.
\newblock {StepCache}: Step-Level Reuse with Lightweight Verification and Selective Patching for {LLM} Serving.
\newblock arXiv:2603.28795.

\bibitem[{Pan et~al.(2025{\natexlab{a}})Pan, Wang, Jia, Karakus, Zancato, Dao, Wang, and Netravali}]{marconi2025}
Pan, R.; Wang, Z.; Jia, Z.; Karakus, C.; Zancato, L.; Dao, T.; Wang, Y.; and Netravali, R. 2025{\natexlab{a}}.
\newblock {Marconi}: Prefix Caching for the Era of Hybrid {LLM}s.
\newblock MLSys 2025 camera-ready version, arXiv:2411.19379.

\bibitem[{Pan et~al.(2025{\natexlab{b}})Pan, Patel, Hu, Shen, Guan, Li, Qin, Wang, and Ding}]{kvflow2025}
Pan, Z.; Patel, A.; Hu, Z.; Shen, Y.; Guan, Y.; Li, W.-L.; Qin, L.; Wang, Y.; and Ding, Y. 2025{\natexlab{b}}.
\newblock {KVFlow}: Efficient Prefix Caching for Accelerating {LLM}-Based Multi-Agent Workflows.
\newblock arXiv:2507.07400.

\bibitem[{Rastogi et~al.(2020)Rastogi, Zang, Sunkara, Gupta, and Khaitan}]{rastogi2020sgd}
Rastogi, A.; Zang, X.; Sunkara, S.; Gupta, R.; and Khaitan, P. 2020.
\newblock Towards Scalable Multi-Domain Conversational Agents: The Schema-Guided Dialogue Dataset.
\newblock In \emph{Proceedings of the AAAI Conference on Artificial Intelligence}, volume~34, 8689--8696.

\bibitem[{Regmi and Pun(2024)}]{gptsemanticcache2024}
Regmi, S.; and Pun, C.~P. 2024.
\newblock {GPT} Semantic Cache: Reducing {LLM} Costs and Latency via Semantic Embedding Caching.
\newblock arXiv:2411.05276.

\bibitem[{R{\"o}ttger et~al.(2024)R{\"o}ttger, Kirk, Vidgen, Attanasio, Bianchi, and Hovy}]{rottger2024xstest}
R{\"o}ttger, P.; Kirk, H.; Vidgen, B.; Attanasio, G.; Bianchi, F.; and Hovy, D. 2024.
\newblock {XSTest}: A Test Suite for Identifying Exaggerated Safety Behaviours in Large Language Models.
\newblock In \emph{Proceedings of the 2024 Conference of the North American Chapter of the Association for Computational Linguistics: Human Language Technologies (Volume 1: Long Papers)}, 5377--5400.

\bibitem[{Ruan et~al.(2026)Ruan, Bi, Zheng, Shi, Wan, and Li}]{cortex2026}
Ruan, C.; Bi, C.; Zheng, K.; Shi, Z.; Wan, X.; and Li, J. 2026.
\newblock {Cortex}: Achieving Low-Latency, Cost-Efficient Remote Data Access for {LLM} via Semantic-Aware Knowledge Caching.
\newblock arXiv:2509.17360.

\bibitem[{Schroeder et~al.(2025)Schroeder, Desai, Cuadron, Chu, Liu, Zhao, Krusche, Kemper, Zaharia, and Gonzalez}]{vcache2025}
Schroeder, L.~G.; Desai, A.; Cuadron, A.; Chu, K.; Liu, S.; Zhao, M.; Krusche, S.; Kemper, A.; Zaharia, M.; and Gonzalez, J.~E. 2025.
\newblock {vCache}: Verified Semantic Prompt Caching.
\newblock Accepted to ICLR 2026, arXiv:2502.03771.

\bibitem[{Shah(2026)}]{groundedcache2026}
Shah, S.~H. 2026.
\newblock Grounded Cache Routing for Retrieval-Augmented Generation: When Is It Safe to Reuse an Answer?
\newblock arXiv:2605.27494.

\bibitem[{Singh et~al.(2026)Singh, Wang, Attaluri, Chiam, and Zhu}]{krites2026}
Singh, A.~K.; Wang, H.; Attaluri, L. N.~S.; Chiam, T.; and Zhu, W. 2026.
\newblock Asynchronous Verified Semantic Caching for Tiered {LLM} Architectures.
\newblock arXiv:2602.13165.

\bibitem[{Wilson(1927)}]{wilson1927probable}
Wilson, E.~B. 1927.
\newblock Probable Inference, the Law of Succession, and Statistical Inference.
\newblock \emph{Journal of the American Statistical Association}, 22(158): 209--212.

\bibitem[{Wu et~al.(2026{\natexlab{a}})Wu, Wang, Zhang, Zhang, Niu, Wu, and Zhang}]{cachepoisoning2026}
Wu, G.; Wang, T.; Zhang, Y.; Zhang, Z.; Niu, J.; Wu, Y.; and Zhang, Y. 2026{\natexlab{a}}.
\newblock When Cache Poisoning Meets {LLM} Systems: Semantic Cache Poisoning and Its Countermeasures.
\newblock In \emph{Proceedings of the Network and Distributed System Security Symposium}.

\bibitem[{Wu et~al.(2026{\natexlab{b}})Wu, Ying, Chen, Gu, and Qu}]{cachemecatchyou2026}
Wu, X.; Ying, L.; Chen, G.; Gu, Y.; and Qu, H. 2026{\natexlab{b}}.
\newblock Cache Me, Catch You: Cache Related Security Threats in {LLM} Serving Frameworks.
\newblock In \emph{Proceedings of the Network and Distributed System Security Symposium}.

\bibitem[{Yan et~al.(2025)Yan, Ni, Chen, Lin, Cheng, Qin, and Ren}]{contextcache2025}
Yan, J.; Ni, W.; Chen, L.; Lin, X.; Cheng, P.; Qin, Z.; and Ren, K. 2025.
\newblock {ContextCache}: Context-Aware Semantic Cache for Multi-Turn Queries in Large Language Models.
\newblock arXiv:2506.22791.

\bibitem[{Yao et~al.(2025)Yao, Li, Liu, Ray, Cheng, Zhang, Du, Lu, and Jiang}]{cacheblend2025}
Yao, J.; Li, H.; Liu, Y.; Ray, S.; Cheng, Y.; Zhang, Q.; Du, K.; Lu, S.; and Jiang, J. 2025.
\newblock {CacheBlend}: Fast Large Language Model Serving for {RAG} with Cached Knowledge Fusion.
\newblock arXiv:2405.16444.

\bibitem[{Yao et~al.(2024)Yao, Shinn, Razavi, and Narasimhan}]{yao2024tau}
Yao, S.; Shinn, N.; Razavi, P.; and Narasimhan, K. 2024.
\newblock {$\tau$}-bench: A Benchmark for Tool-Agent-User Interaction in Real-World Domains.
\newblock arXiv:2406.12045.

\bibitem[{Zhang and Choi(2021)}]{zhang2021situatedqa}
Zhang, M.; and Choi, E. 2021.
\newblock {SituatedQA}: Incorporating Extra-Linguistic Contexts into {QA}.
\newblock In \emph{Proceedings of the 2021 Conference on Empirical Methods in Natural Language Processing}, 7371--7387.

\bibitem[{Zhang et~al.(2026)Zhang, Liu, Xie, Huang, and She}]{cacheattack2026}
Zhang, Z.; Liu, Z.; Xie, Y.; Huang, Q.; and She, D. 2026.
\newblock From Similarity to Vulnerability: Key Collision Attack on {LLM} Semantic Caching.
\newblock Accepted to ICML 2026, arXiv:2601.23088.

\bibitem[{Zheng et~al.(2024)Zheng, Yin, Xie, Sun, Huang, Yu, Cao, Kozyrakis, Stoica, Gonzalez, Barrett, and Sheng}]{sglang2024}
Zheng, L.; Yin, L.; Xie, Z.; Sun, C.; Huang, J.; Yu, C.~H.; Cao, S.; Kozyrakis, C.; Stoica, I.; Gonzalez, J.~E.; Barrett, C.; and Sheng, Y. 2024.
\newblock {SGLang}: Efficient Execution of Structured Language Model Programs.
\newblock arXiv:2312.07104.

\bibitem[{{Zilliz}(2023)}]{gptcache2023}
{Zilliz}. 2023.
\newblock {GPTCache}: A Library for Creating Semantic Cache for LLM Queries.
\newblock \url{https://github.com/zilliztech/GPTCache}.
\newblock Accessed 2026-07-04.

\end{thebibliography}
